# Gradient Sparsification for Communication-Efficient Distributed Optimization

Jianqiao Wangni*, Jialei Wang‡, Ji Liu†♯, and Tong Zhang*

*Tencent AI Lab, Shenzhen, China
†Tencent AI Lab, Seattle, USA
‡Department of Computer Science, University of Chicago, IL, USA
♯Department of Computer Science, University of Rochester, NY, USA

**Abstract**

Modern large scale machine learning applications require stochastic optimization algorithms to be implemented on distributed computational architectures. A key bottleneck is the communication overhead for exchanging information such as stochastic gradients among different workers. In this paper, to reduce the communication cost we propose a convex optimization formulation to minimize the coding length of stochastic gradients. To solve the optimal sparsification efficiently, several simple and fast algorithms are proposed for approximate solution, with theoretical guaranteed for sparseness. Experiments on $\ell_2$ regularized logistic regression, support vector machines, and convolutional neural networks validate our sparsification approaches.

## 1 Introduction

Modern large scale machine learning applications require scaling stochastic optimization algorithms to distributed computational architectures. A key bottleneck is the communication overhead for exchanging information among different workers. For example, we have $n$ training data distributed on $M$ workers, and each of them owns its local copy of the model parameter vector. In the synchronized stochastic gradient method, each worker processes a random minibatch of its training data, and then the local updates are synchronized by making an *All-Reduce* step, which aggregates stochastic gradients from all workers, and taking a *Broadcast* step that transmits the updated parameter vector back to all workers. The process is repeated until an appropriate convergence criterion is met. An important factor that may significantly slow down any optimization algorithm is the communication cost among workers. Even for the single machine multi-core setting, where the cores communicate with each other by reading and writing to a chunk of shared memory, conflicts of (memory access) resources may significantly degrade the efficiency.

The existing work on distributed machine learning mainly focus on how to design communication efficient algorithms to reduce the round of communications among workers, such as [33, 23, 19]. More recently, several papers considered the problem of reducing the precision of gradient by using fewer bits to represent floating pointing numbers [28, 1, 30]. In this paper we propose a novel approach to complement these methods above. Specifically, we sparsify stochastic gradients appropriately to reduce the communication cost, with minor sacrifice on the number of iterations.

The key idea behind of our sparsification technology is to drop some coordinates of the stochastic gradient and appropriately amplify the remaining coordinates to ensure the unbiasness of the sparsified stochastic gradient. The sparsification approach can significantly reduce the coding length of the stochastic gradient and only lightly increase the variance of the stochastic gradient. This paper proposes a convex formulation to achieve the best tradeoff of variance and sparsity: the optimal probabilities to sample coordinates can be obtained given any fixed variance budget. To solve this optimization in a linear time, several efficient algorithms are proposed to find approximate optimal solutions with sparsity guarantees.

The proposed sparsification approach can be encapsulated seamlessly to many bench-mark stochastic optimization algorithms in machine learning, such as SGD, SVRG, and ADAM [31, 2, 8, 9]. We provide experiments to validate the proposed approach using $\ell_2$ regularized logistic regression, support vector machines, and convolutional neural networks on both synthetic and real date sets.

## 2 Related works

The problem of communication-efficient large scale optimization has drawn significant attention in recent years. It is known that the communication complexity for optimizing a strongly-convex and smooth function is proved to be at least $O(d\log(d) + d\log(1/\epsilon))$ for an $\epsilon$-approximate solution [24].

In modern applications, stochastic first order methods such as stochastic gradient descent (SGD) methods are preferred, and they achieve better computational complexities than gradient descent (GD), both for practical convex and nonconvex problems. The vanilla SGD can only yield a solution with a sublinear convergence rate. For strongly convex problems, variance reduction methods such as stochastic dual coordinate ascent (SDCA) [22], stochastic average gradient (SAG) [20], stochastic variance-reduce gradient (SVRG) [8] and SAGA [6], can lead to linear convergence rate.

In recent years, a number of researchers proposed communication-efficient distributed optimization algorithms. The one-shot averaging algorithm [33] simply averages the solutions obtained by all machines, and proved that this is statistical optimal under relatively strong assumptions. To remedy its limitation, [23] proposed distributed approximate newton algorithm (DANE), which tries to solve a sub-sampled newton iteration with full first order gradients at each round, and the computation of full first order gradient require communication. Its improved version, AIDE [19], is more practical for implementation since it only calculates the Newton direction inexactly. The DiSCo algorithm [32] also employs inexact Newton steps, where the Newton direction is obtained by solving a linear system using a preconditioned conjugate gradient algorithm. The authors further showed that the number of communication round of the algorithm grows very slowly as the number of machines increases, for self-concordant loss functions.

In practice, one often employs SGD type of algorithms for large scale machine learning, and a number of researchers tried to address the communication problem when we directly parallelize SGD. Although it is well known that SGD converges more slowly with a large mini-batch size, by iteratively solving a modified problem that depends on the large mini-batches, SGD is able to converge at about the same speed as that of the small mini-batch counterpart [14, 26, 27]. Moreover, the asynchronous version of SVRG [18, 34] can achieve near linear speed up under appropriate assumptions, and can be easily deployed on large clusters. Practice implementation of distributed machine often employs a parameter server, such as [7, 13, 12, 29], where the worker nodes pull parameters from the parameter server periodically, calculating the local gradients and push them back to the server, who further update the global copy. Under the staleness synchronous protocol, which requires a bounded maximum delay between the fastest and the slowest machines, the algorithm is assured to converge.

A different technique is to reduce the precision of gradients, investigated in the literature recently. It can reduce the communications cost over the network. The low-bit SGD algorithm [4] considers the energy efficiency from both software and hardware perspectives. In particular, the 1Bit-SGD [21], as a more aggressive compression approach, only transmits the sign of each coordinate of gradients. This heuristic is also effective in training convolutional neural networks, as in dorefa-net[35], which compresses the bit width of weights, activations, and gradients of neural nets to 1,2,and 6, respectively. Their experiments on AlexNet [10] showed only a slight drop in accuracy. Similarly, the ternary gradient [28] approach compresses each coordinate to three numerical levels $\{-1, 0, +1\}$. The quantized SGD algorithm [1] further compresses the gradients into multiple numerical precisions but without introducing any bias, which is achieved by a random rounding scheme. Their experiments on neural networks showed an improved training speed.

# 3 Algorithms

We consider the problem of sparsifying a stochastic gradient vector, and formulate it as a linear planning problem. The following notations will be used throughout the paper. Consider a training data set $\{x_n\}_{n=1}^N \subset \mathbb{R}^d$, and each training data point $x_n$ is associating with a loss function $f_n : \mathbb{R}^d \to \mathbb{R}$, that is associating with the $n$th data point $x_n$. We use $w \in \mathbb{R}^d$ to denote the model parameter vector, and consider solving the following optimization problem:

$$\min_w \quad f(w) := \frac{1}{N}\sum_{n=1}^{N} f_n(w). \tag{1}$$

Stochastic optimization methods typically employ an unbiased stochastic gradient $g_t(w)$ at each time $t$ that satisfies $\mathbb{E}\,[g_t(w)] = \nabla f(w)$. A commonly used update for solving (1) is

$$w_{t+1} = w_t - \eta_t g_t(w_t),$$

where $g_t(w_t)$ is an unbiased estimate for the true gradient $\nabla f(w_t)$. *The above derivation implies that the convergence of SGD is significantly dominated by $\mathbb{E}\|g_t(w_t)\|^2$ or equivalently the variance of $g_t(w_t)$.* It can be seen from the following simple derivation. Assume that the loss function $f(w)$ is $L$-smooth with respect to $w$, which means that for $\forall x, y \in \mathbb{R}^d$, $\|\nabla f(x) - \nabla f(y)\| \le L\|x - y\|$ (where $\|\cdot\|$ is the $\ell_2$-norm). Then the expected loss function is given by

$$\begin{aligned}
\mathbb{E}\,[f(w_{t+1})] \le & \mathbb{E}\left[f(w_t) + \nabla f(w_t)^\top (x_{t+1} - x_t) + \frac{L}{2}\|x_{t+1} - x_t\|^2\right] \\
= & \mathbb{E}\,\left[f(w_t) - \eta_t \nabla f(w_t)^T g_t(w_t) + \frac{L}{2}\eta_t^2\|g_t(w_t)\|^2\right] \\
= & f(w_t) - \eta_t\|\nabla f(w_t)\|^2 + \frac{L}{2}\eta_t^2 \underbrace{\mathbb{E}\,\|g_t(w_t)\|^2}_{\text{variance}},
\end{aligned}$$

where the inequality is due to the Lipschitzian property. The first equality is due to (2), and the second equality is due to the unbiased nature of the gradient $\mathbb{E}\,[g_t(w)] = \nabla f(w)$. So the magnitude of $\mathbb{E}(\|g_t(w_t)\|^2)$ or equivalently the variance of $g_t(w_t)$ will significantly affect the convergence efficiency. The following are two popular ways to choose $g_t(w_t)$

- SGD [31, 2, 25]

$$g_t(w_t) = \nabla f_{n_t}(w_t) \tag{2}$$

where $n_t$ is uniformly sampled from the data set;

- SVRG [8]

$$g_t(w_t) = \nabla f_{n_t}(w_t) - \nabla f_{n_t}(\widetilde{w}) + \nabla f(\widetilde{w}) \tag{3}$$

where $n_t$ is uniformly sampled from the data set and $\widetilde{w}$ is a reference point.

Next we consider how to reduce the communication cost in distributed machine learning by using a sparsification of stochastic gradient $g_t(w_t)$, denoted by $Q(g(w_t))$, such that $Q(g_t(w_t))$ is unbiased, and has a relatively small variance. In the following, to simplify notation, we denote the current stochastic gradient $g_t(w_t)$ by $g$ for short, in which we drop the subscript $t$ and $w_t$. Note that $g$ can be obtained either by SGD or SVRG. We also let $g_i$ be the $i$-th component of vector $g \in \mathbb{R}^d$: $g = [g_1, \dots, g_d]$. We propose to randomly drop out the $i$-th coordinate by a probability of $1 - p_i$, which means that the coordinates remains non-zero with a probability of $p_i$. Let $Z_i \in \{0, 1\}$ be a binary-valued random variable indicating whether the $i$-th coordinate is selected: $Z_i = 1$ with probability $p_i$ and $Z_i = 0$ with probability $1 - p_i$. Then, to make the resulting sparsified gradient vector $Q(g)$ unbiased, we amplify the non-zero coordinates, from $g_i$ to $g_i/p_i$. So the final sparsified vector is $Q(g)_i = Z_i(g_i/p_i)$. The whole protocol can be summarized as follows:

$$\begin{aligned}
\text{original vector } g = \quad & [g_1, g_2, \cdots, g_d], \\
\text{probability vector } p = \quad & [p_1, p_2, \cdots, p_d], \\
\text{selection vector } Z = \quad & [Z_1, Z_2, \cdots, Z_d], \qquad \text{where } P(Z_i = 1) = p_i \\
\text{sparsified vector } Q(g) = \quad & \left[ Z_1 \frac{g_1}{p_1}, Z_2 \frac{g_2}{p_2}, \cdots, Z_d \frac{g_d}{p_d} \right]
\end{aligned}$$

We note that if $g$ is an unbiased estimate of the gradient, then $Q(g)$ is also an unbiased estimate of the gradient since

$$\mathbb{E}\left[Q(g)_i\right] = p_i \times \frac{g_i}{p_i} + (1 - p_i) \times 0 = g_i.$$

In distributed machine learning, each worker calculates gradient $g$ and transmits it to the master node or the parameter server for update. We use an index $m$ to indicate a node, and assume there are totally $M$ nodes. The gradient sparsification method can be used with a synchronous distributed stochastic optimization algorithm in Algorithm 1. Asynchronous algorithms can also be used with our technique in a similar fashion.

### 3.1 Mathematical formulation

Although the gradient sparsification technique can reduce communication cost, it increases the variance of the gradient vector, which might slow down the convergence rate. In the following section we will investigate how to find the best tradeoff between sparsity and variance for the sparsification technique. In particular, we consider given a budget of maximal variance, how to

**Algorithm 1** A synchronous distributed optimization algorithm

Initialize the clock $t = 0$ and initialize the weight $w_0$.
**repeat**
    Each worker $m$ calculates stochastic gradient $g^m(w_t)$ using local data as in Eq. (2) or (3).
    Calculate the probability vector $p^m$.
    Sparsify the gradients to $Q(g^m(w_t))$.
    Take an *All-Reduce* step to obtain an averaged gradient $v_t = \frac{1}{M}\sum_{m=1}^{M} Q(g^m(w_t))$.
    (Optional) sparsify the averaged gradient as $v_t = Q(v_t)$
    Broadcast the average gradient $v_t$ to all workers
    Take a descent step $w_{t+1} = w_t - \eta_t v_t$ on all workers
    Update the clock $t = t + 1$.
**until** convergence

find out the optimal sparsification strategy. First note that the variance of $Q(g)$ can be bounded by

$$\mathbb{E}\sum_{i=1}^{d}[Q(g)_i^2] = \sum_{i=1}^{d}\left[\frac{g_i^2}{p_i^2} \times p_i + 0 \times (1-p_i)\right] = \sum_{i=1}^{d}\frac{g_i^2}{p_i}.$$

In addition, the expected sparsity of $Q(g_i)$ is given by

$$\mathbb{E}\left[\|Q(g)\|_0\right] = \sum_{i=1}^{d} p_i.$$

In this section, we try to balance these two factors (sparsity and variance) by formulating it as a linear planning problem as follows:

$$\begin{aligned} \min_{p} \quad & \sum_{i=1}^{d} p_i \\ \text{s.t.} \quad & \sum_{i=1}^{d}\frac{g_i^2}{p_i} \le (1+\epsilon)\sum_{i=1}^{d} g_i^2, \qquad 0 < p_i \le 1 \quad \forall i \in [d], \end{aligned} \tag{4}$$

where $\epsilon$ is a factor that controls the variance increase of the stochastic gradient $g$.

This leads to an optimal strategy for sparsification given an upper bound on the variance. The following proposition provides a closed-form solution for problem (4).

**Proposition 1.** *The solution of the optimal sparsification problem* (4) *is given by a probability vector $p$ such that*

$$p_i = \min(\lambda|g_i|, 1), \qquad \forall i \in [d],$$

*where $\lambda > 0$ is a certain independent constant only depend on $g$ and $\epsilon$.*

*Proof.* By introducing Lagrange multipliers $\lambda$ and $\mu_i$, we know that the solution of (4) is given by the solution of the following objective:

$$\min_{p}\max_{\lambda}\max_{\mu} L(p_i, \lambda, \mu_i) = \sum_{i=1}^{d} p_i + \lambda^2\left(\sum_{i=1}^{d}\frac{g_i^2}{p_i} - (1+\epsilon)\sum_{i=1}^{d} g_i^2\right) + \sum_{i=1}^{d}\mu_i(p_i - 1).$$

Consider the KKT conditions of the above formulation, by stationarity with respect to $p_i$ we have:

$$1 - \lambda^2 \frac{g_i^2}{p_i^2} + \mu_i = 0, \quad \forall i \in [d].$$

Combined with the complementary slackness condition that guarantees $\mu_i(p_i - 1) = 0, \forall i \in [d]$, we obtain the following connections:

$$p_i = \begin{cases} 1, & \text{if } \mu_i \neq 0 \\ \lambda |g_i|, & \text{if } \mu_i = 0. \end{cases}$$

Above formula tells us that for several coordinates the probability of keeping the value is 1 (when $\mu_i \neq 0$), and for other coordinates the probability of keeping the value is proportional to the magnitude of the gradient $g_i$. Also, by simple reasoning we know that if $|g_i| \geq |g_j|$ then $|p_i| \geq |p_j|$ (otherwise we simply switch $p_i$ and $p_j$). Therefore there is a dominating set of coordinates $S$ with $p_j = 1, \forall j \in S$, and it must be the set of $|g_j|$ with the largest absolute magnitudes. Suppose this set has a size $k$ ($0 \leq k \leq d$) and denote by $g_{(1)}, g_{(2)}, ..., g_{(d)}$ the components of $g$ ordered by their magnitudes (for the largest to the smallest), we have

$$p_{(i)} = \begin{cases} 1, & \text{if } i \leq k \\ \lambda |g_{(i)}|, & \text{if } i > k. \end{cases} \tag{5}$$

It implies our claim. □

### 3.2 Sparsification algorithms

In this section we propose two algorithms for efficiently calculating the optimal probability vector $p$ in Proposition 1. Since $\lambda > 0$, by complementary slackness condition we have

$$\sum_{i=1}^{d} \frac{g_i^2}{p_i} - (1+\epsilon) \sum_{i=1}^{d} g_i^2 = \sum_{i=1}^{k} g_{(k)}^2 + \sum_{i=k+1}^{d} \frac{|g_{(i)}|}{\lambda} - (1+\epsilon) \sum_{i=1}^{d} g_i^2 = 0,$$

which further implies

$$\lambda = \frac{\sum_{i=k+1}^{d} |g_{(i)}|}{\epsilon \sum_{i=1}^{d} g_i^2 + \sum_{i=k+1}^{d} g_{(i)}^2}.$$

Using the constraint $\lambda |g_{(k+1)}| \leq 1$, we have

$$|g_{(k+1)}| \left( \sum_{i=k+1}^{d} |g_{(i)}| \right) \leq \epsilon \sum_{i=1}^{d} g_i^2 + \sum_{i=k+1}^{d} g_{(i)}^2.$$

It follows that we should find the smallest $k$ which satisfies the above inequality. Based on above reasoning, we get the following closed-form solution for $p_i$ in Algorithm 2.

In practice, using Algorithm 2 to find $S_k$ requires partial sorting of the gradient magnitude values, which could be computationally expensive. Therefore we developed a greedy algorithm for approximately solving the problem. We pre-define a sparsity parameter $\rho \in [0, 1]$, which implies we aim to find $p_i$ that satisfies $\sum_i p_i / d \approx \rho$. Loosely speaking, we want to initially set $\widetilde{p}_i =$

**Algorithm 2** Closed Form Solution

Find the smallest $k$ such that

$$|g_{(k+1)}| \left( \sum_{i=k+1}^{d} |g_{(i)}| \right) \le \epsilon \sum_{i=1}^{d} g_i^2 + \sum_{i=k+1}^{d} g_{(i)}^2, \tag{6}$$

is true, and let $S_k$ be the set of coordinates with top $k$ largest magnitude of $|g_i|$.

Set the probability vector $p$ by

$$p_i = \begin{cases} 1, & \text{if } i \in S_k \\ \frac{|g_i|(\sum_{j=k+1}^{d} |g_{(j)}|)}{\epsilon \sum_{j=1}^{d} g_{(j)}^2 + \sum_{j=k+1}^{d} g_{(j)}^2}, & \text{if } i \notin S_k. \end{cases} \tag{7}$$

$\rho d |g_i| / \sum_i |g_i|$, which sums to $\sum_i \widetilde{p}_i = \rho d$, meeting our requirement on $\rho$. However, by the truncation operation $p_i = \min(\widetilde{p}_i, 1)$, the expected nonzero density will be less than $\rho$. Now, we can use an iterative procedure, where in the next iteration, we fix the set of $\{p_i : p_i = 1\}$ and scale the remaining values, as summarized in Algorithm 3.

The algorithm is much easier to implement, and computationally more efficient. Since the operations mainly consist of accumulations, multiplications and minimizations, they can be easily accelerated on hardware supporting *single instruction multiple data (SIMD)*, including modern Intel CPUs with *SSE/AVX* instructions and ARM CPUs with *NEON* instructions.

**Algorithm 3** Greedy Algorithm

**Input** $g \in \mathbb{R}^d$, $\rho \in [0, 1]$
Initialize $p^0 \in \mathbb{R}^d$, $j = 0$.
Set $p_i^0 = \min\left(\rho d |g_i| / \sum_i |g_i|, 1\right)$ for all $i$.
**repeat**
  Identify the active set $\mathcal{I} = \{1 \le i \le D | p_i^j \neq 1\}$.
  Compute the scaling variable $c = (\rho d - d + |\mathcal{I}|) / \sum_{i \in \mathcal{I}} p_i^j$.
  If $c \le 1$, break the loop.
  Recalibrate the values by $p_i^{j+1} = \min(c p_i^j, 1)$.
  $j = j + 1$
**until** convergence
return $p = p^j$

### 3.3 Coding strategy

Once we have computed a sparsified gradient vector $Q(g)$, we need to pack the resulting vector into a message for transmission. Here we apply a hybrid strategy for coding $Q(g)$. Suppose that the computer represents a floating point scalar using $b$ bits, which is enough for a precise representation of any variables with negligible loss in precision. We use two vectors for representing non-zero coordinates, one for coordinates $i \in S_k$, and the other for coordinates $i \notin S_k$. The vector $Q_A(g)$ represents $\{g_i : i \in S_k\}$, where each item of $Q_A(g)$ needs $\log d$ bits to represent the coordinates and $b$ bits for the value $g_i/p_i$. The vector $Q_B(g)$ represents $\{g_i : i \notin S_k\}$, since in this case, we have $p_i = \lambda|g_i|$, we have for all $i \notin S_k$ the quantized value $Q(g_i) = g_i/p_i = \text{sign}(g_i)/\lambda$. Therefore to represent $Q_B(g)$, we only need one floating point $1/\lambda$, plus the non-zero coordinates $i$ and its sign $\text{sign}(g_i)$. Here we give an example about the format,

$$\text{sparsified vector}: \quad \left[\frac{g_1}{p_1}, 0, 0, \frac{g_4}{p_4}, \frac{g_5}{p_5}, \frac{g_6}{p_6}, \cdots, 0\right], \tag{8}$$

$$\text{Vector } Q_A(g): \quad \left[1, \frac{g_1}{p_1}, 5, \frac{g_5}{p_5} \cdots, 0\right], \qquad (i = 1, i = 5 \in S_k) \tag{9}$$

$$\text{Vector } Q_B(g): \quad [4, -1/\lambda, 6, 1/\lambda, \cdots]. \qquad (i = 4, i = 6 \notin S_k, g_4 < 0, g_6 > 0) \tag{10}$$

Moreover, we can also represent the indices of $A$ and vector $Q_B(g)$ using a dense vector of $\widetilde{q} \in \{0, \pm 1, 2\}^d$, where each component $\widetilde{q}_i$ is defined as $Q(g_i) = \lambda Q(g_i)$ when $i \notin S_k$ and $\widetilde{q}_i = 2$ if $i \in S_k$. Using the standard entropy coding, we know that $\widetilde{q}$ requires at most $\sum_{\ell=-1}^{2} d_\ell \log_2(d/d_\ell) \leq 2d$ bits to represent.

# 4 Theoretical guarantees on sparsity

In this section we analyze the expected sparsity of $Q(g)$, which equals to $\sum_{i=1}^d p_i$. In particular we show when the distribution of gradient magnitude values is highly skewed, there is a significant gain in applying the proposed sparsification strategy. First, we define the following notion of approximate sparsity on the magnitude at each coordinate of $g$:

**Definition 2.** *A vector $g \in \mathbb{R}^d$ is $(\rho, s)$-approximately sparse if there exists a subset $S \subset [d]$ such that $|S| = s$ and*

$$\|g_{S^c}\|_1 \leq \rho \, \|g_S\|_1 \,, \tag{11}$$

*where $S^c$ is the complement of $S$.*

The notion of $(\rho, s)$-approximately sparsity is inspired by the restricted eigenvalue condition used in high-dimensional statistics [3]. $(\rho, s)$-approximately sparsity measures how well the signal of a vector is concentrated on a small subset of the coordinates of size $s$. As we will see later, the quantity $(1 + \rho)s$ plays an important role in establish the expected sparsity bound. Note that we can always take $s = d$ and $\rho = 0$ so that $(\rho, s)$ satisfies the above definition with $(1+\rho)s \leq d$. If the distribution of magnitude values in $g$ is highly skewed, we would expect the existence of $(\rho, s)$ such that $(1 + \rho)s \ll d$. For example when $g$ is exactly $s$-sparse, we can choose $\rho = 0$ and the quantity $(1 + \rho)s$ reduces to $s$ which can be significantly smaller than $d$.

**Lemma 3.** *If the gradient $g \in \mathbb{R}^d$ of the loss function is $(\rho, s)$-approximately sparse as in Definition 2. Then we can find a sparsification $Q(g)$ with $\epsilon = \rho$ in (4) (that is, the variance of $Q(g)$ is increased by a factor of no more than $1+\rho$), and the expected sparsity of $Q(g)$ can be upper bounded by*

$$\mathbb{E}\left[\|Q(g)\|_0\right] \leq (1 + \rho)s. \tag{12}$$

*Proof.* Based on Definition 2, we can choose $\epsilon = \rho$ and $S_k = S$ that satisfies (6), thus

$$\begin{aligned}
\mathbb{E}\left[\|Q(g)\|_0\right] =& \sum_{i=1}^{d} p_i = \sum_{i \in S_k} p_i + \sum_{i \notin S_k} p_i \\
=& s + \sum_{i \notin S_k} \frac{|g_i|(\sum_{j=k+1}^{d} |g_{(j)}|)}{\epsilon \sum_{j=1}^{k} g_{(j)}^2 + (1+\epsilon)\sum_{j=k+1}^{d} g_{(j)}^2} \\
=& s + \frac{\left\|g_{S_k^c}\right\|_1^2}{\rho \left\|g_{S_k}\right\|_2^2 + (1+\rho)\left\|g_{S_k^c}\right\|_2^2} \\
\leq& s + \frac{\rho^2 s \left\|g_{S_k}\right\|_2^2}{\rho \left\|g_{S_k}\right\|_2^2 + (1+\rho)\left\|g_{S_k^c}\right\|_2^2} \\
\leq& (1+\rho)s,
\end{aligned}$$

which completes the proof. □

**Remark 1.** *Lemma 3 indicates that the variance after sparsification only increase by a factor of $(1+\rho)$, while in expectation we only need to communicate a $(1+\rho)s$-sparse vector after sparsified. In order to achieve the same optimization accuracy, we may need to increase the number of iterations by a factor up to $(1+\rho)$, and the overall number of floating point numbers communicated is reduced by a factor of up to $(1+\rho)^2 s/d$.*

Above lemma shows the number of floating point numbers needed to communicate is reduced by the proposed sparsification strategy. As shown in Section 3.3, we only need to use one floating point number to encoding the gradient values in $S_k^c$, so there is a further reduction in communication when considering the total number of bits transmitted, this is characterized by the Theorem below.

**Theorem 4.** *If the gradient $g \in \mathbb{R}^d$ of the loss function is $(\rho, s)$-approximately sparse as in Definition 2, and a floating point scalar costs $b$ bits, then the coding length of $Q(g)$ in Lemma 3 can be bounded by $s(b + \log_2 d) + \min(\rho s \log_2 d, d) + b$.*

*Proof.* The proof is an extension of Lemma 3. We use $\mathcal{H}$ to represent the coding length. Then

$$\begin{aligned}
\mathbb{E}\left[\mathcal{H}[Q(g)]\right] =& \mathbb{E}\left[\mathcal{H}[Q_A(g)]\right] + \mathbb{E}\left[\mathcal{H}[Q_B(g)]\right] \qquad (13) \\
=& \sum_{i \in S_k} p_i(b + \log_2 d) + \min\left(d, \sum_{i \notin S_k} p_i \log_2 d\right) + b \\
\leq& s(b + \log_2 d) + \min\left(d, \frac{\rho^2 s \left\|g_{S_k}\right\|_2^2}{\rho \left\|g_{S_k}\right\|_2^2 + (1+\rho)\left\|g_{S_k^c}\right\|_2^2} \log_2 d\right) + b \\
\leq& s(b + \log_2 d) + \min(\rho s \log_2 d, d) + b.
\end{aligned}$$

Here the last term $b$ indicates the coding length of $\lambda$, $|S_k^c|$ indicates the space complexity of signs of vector $Q_B(g)$, and the min operator minimize the coding length of vector $Q_B(g)$ over two strategies. □

The coding length of the original gradient vector $g$ is $db$, by considering the slightly increased number of iterations to reach the same optimization accuracy, the total communication cost is reduced by a factor of at least $(1+\rho)((s+1)b + \log_2 d)/db$.

## 5 Experiments

In this section we conduct experiments to validate the effectiveness and efficiency of the proposed sparsification technique. We use $\ell_2$ regularized logistic regression as an example for convex problems, and take convolutional neural networks as an example for non-convex problems. The sparsification technique show strong improvement over a baseline of uniform sampling approach, the iteration complexity is relatively less increased comparing to the communication costs we saved. Moreover, we also conduct asynchronous parallel experiments on the shared memory architecture. In particular, our experiments show that the proposed sparsification technique significantly reduces the conflict among multiple threads and dramatically improves the performance.

In all experiments, the probability vector $p$ is calculated by Algorithm 3. In practice, we find that the greedy algorithmis able to produce a high quality approximation of the optimal $p$ vector, after the two iteration ($j = 2$ and $p \leftarrow p^j$), for convex problems and for neural networks, since the further update of $p^{j+1} - p^j$ is comparably negligible to $p^j$ by more than one order.

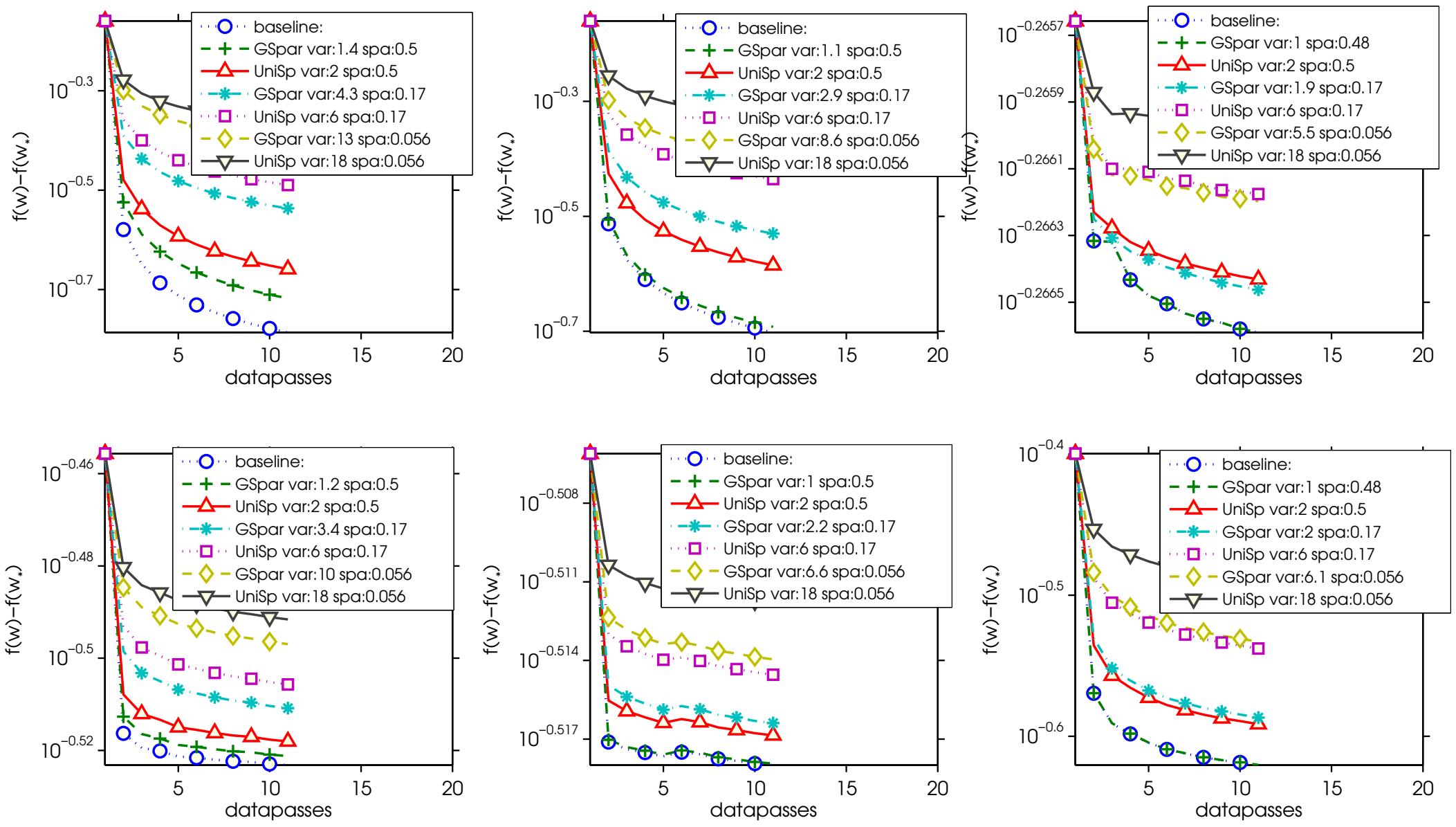

Figure 1: SGD . Datasets generated by setting $C_1 = 0.6$. (Weaker sparsity)

### 5.1 Experiments on convex problems

We first validate the sparsification technique on the $\ell_2$ regularized logistic regression problem using SGD and SVRG respectively:

$$f(w) = \frac{1}{N}\sum_n \log_2\left(1 + \exp(-a_n^\top w b_n)\right) + \lambda_2 \|w\|_2^2, \quad a_n \in \mathbb{R}^d, \quad b_n \in \{-1, 1\}. \tag{14}$$

The experiments are conducted on synthetic data for the convenience to control the data sparsity. The algorithm is implemented using MATLAB. We implement the gradient-sparsified SGD with a diminishing step size $\eta_t \propto 1/(t \cdot var)$, where the $var$ is a calculated result as $var =$ $||Q\left[g(w_t)\right]||^2/||g(w_t)||^2$. This modification over the typical SGD step size of $\eta_t \propto 1/t$ can be inferred from the convergence analysis. We set gradient-sparsified SVRG with a constant step size divided by the variance factor, as $\eta_t \propto 1/var$, as a modification of the constant step size of SVRG. The mini-batch size is set to be 8 by default unless specified. We simulated with $M = 4$ machines, where one machine is both a worker and the master that aggregates stochastic gradients received from other workers. We compare our algorithm with a uniform sampling method as baseline, where each element of the probability vector is set to be $p_i = \rho$. In this method, the sparsified vector is with the sparsity ratio $\rho$ in expectation.

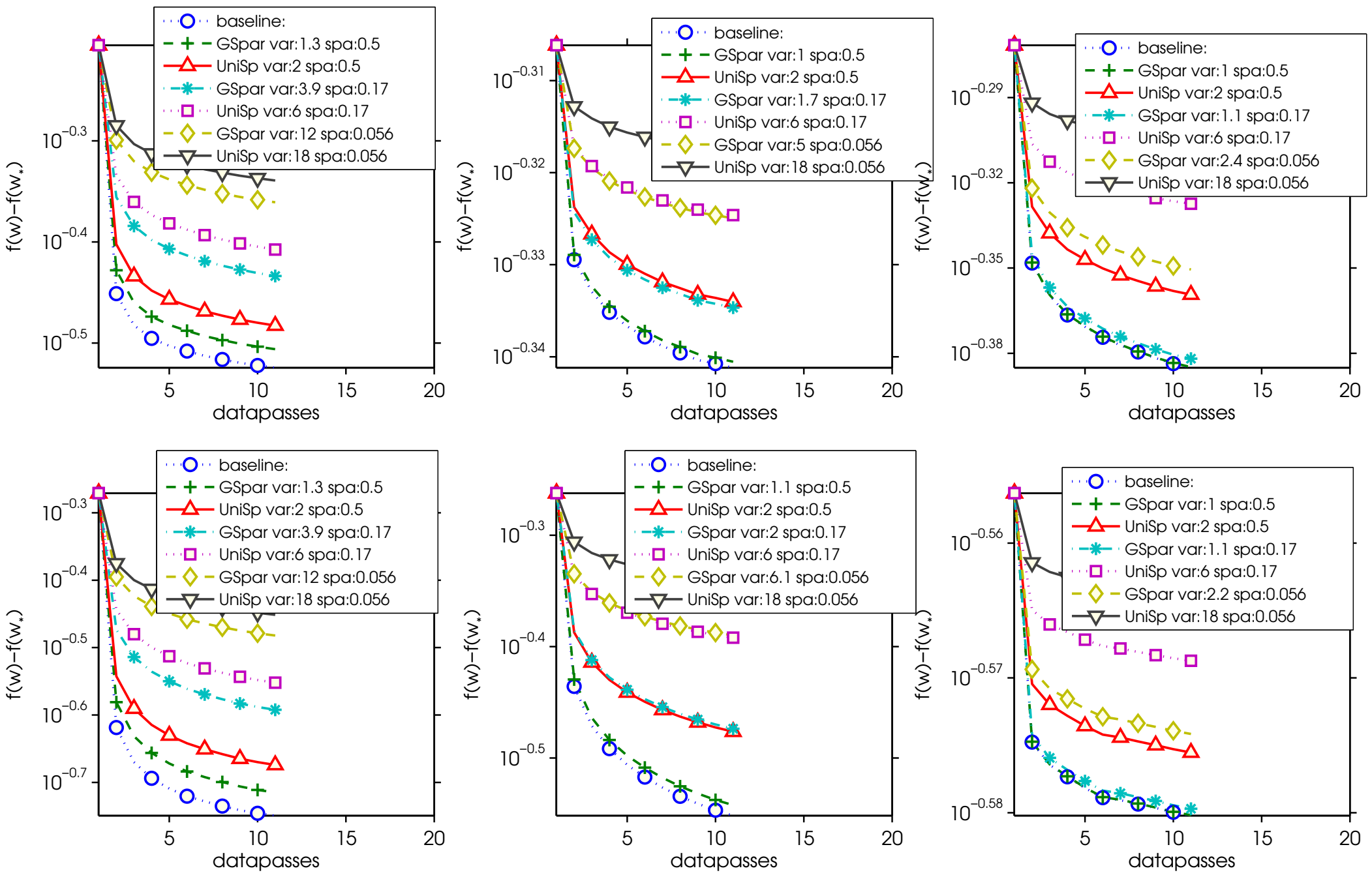

Figure 2: SGD . Datasets generated by setting $C_1 = 0.9$. (Stronger sparsity)

For SGD, the sparsification of stochastic gradient is implemented using Algorithm 1. For SVRG, there are two possible application of the proposed sparsification technique.

- The sparsification procedure can be put on the variance reduced gradient, and the workers transmit $Q(g_t(w_t) - g_t(\widetilde{w}) + \nabla f(\widetilde{w}))$ to the master node.
- The second choice is to keep an accurate full gradient $\nabla f(\widetilde{w})$ in the master node, and this only costs one round of communication after the update of the reference vector $\widetilde{w}$, each worker calculates the sparsified local gradients $Q(g^m(w_t) - g^m(\widetilde{w}))$, then the master node take a *All-Reduce* step and *Broadcast* the updated weight

$$w_{t+1} = w_t - \eta\left(\nabla f(\widetilde{w}) + \frac{1}{M}\sum_m Q\left(g^m(w_t) - g^m(\widetilde{w})\right)\right) \tag{15}$$

to workers.

We found that in the experiments, each implementation has advantages under different settings, which may probably due to the reason that they show no obviously different convergence rate in theoretical analysis. So there is no general conclusion that which one should be used, so the programmer may choose one implementation by testing. To keep the notations in accordance with ones of SGD, and to show the relation between the sparsity, gradient variance and convergence, we use the first implementation for following experiments.

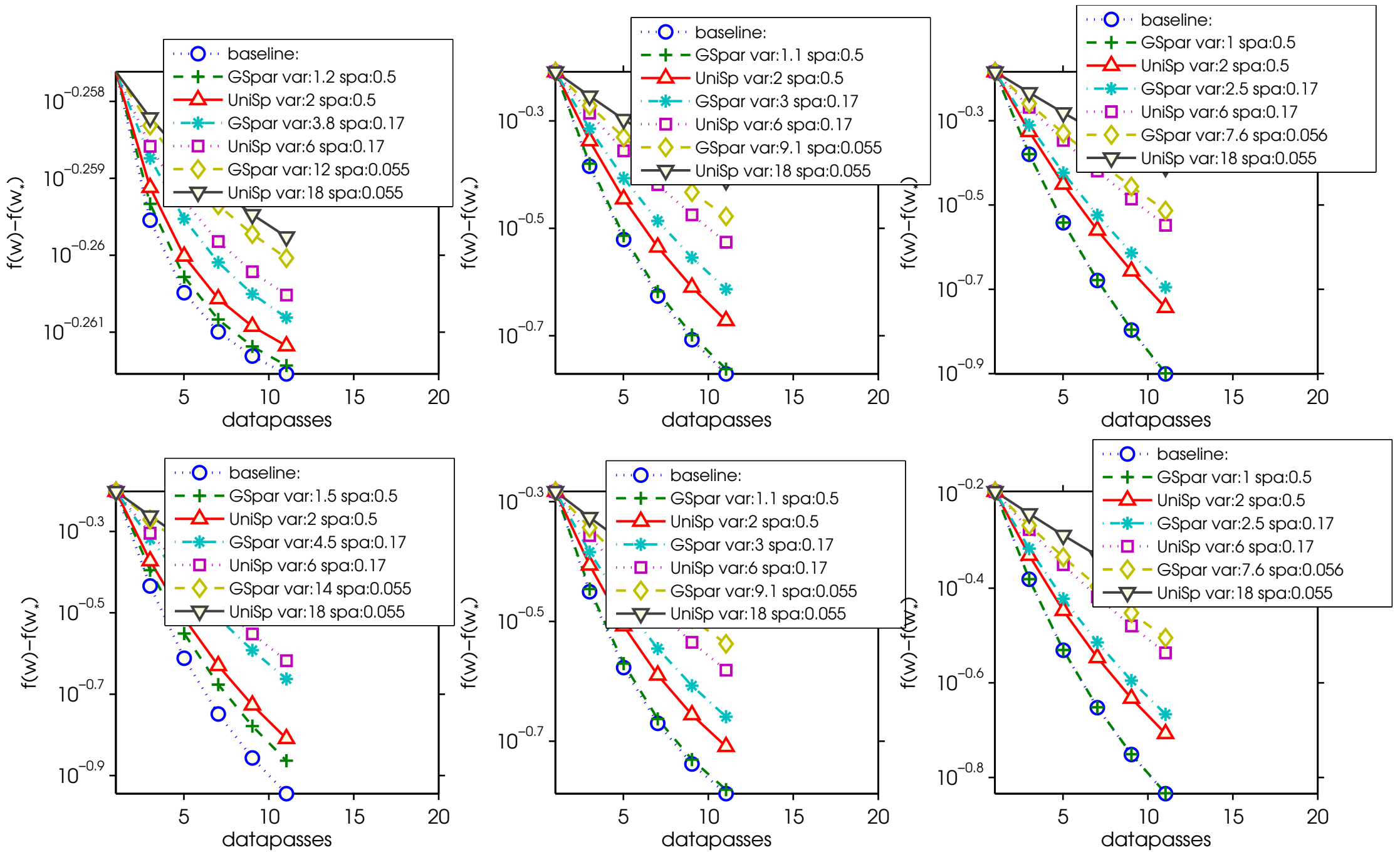

Figure 3: SVRG. Datasets generated by setting $C_1 = 0.6$. (Weaker sparsity)

The data set $\{x_n\}_{n=1}^N$ is generated as follows

$$\begin{aligned}
\text{dense data generation:} \quad & \bar{x}_{ni} \sim \mathcal{N}(0,1), \quad \forall i \in [d], n \in [N], \\
\text{magnitude sparsification:} \quad & \bar{B} \sim \text{Uniform}[0,1]^d, \quad \bar{B}_i \leftarrow C_1 \bar{B}_i, \quad \text{if:} \bar{B}_i \leq C_2, \quad \forall i \in [d] \\
\text{data sparsification:} \quad & x_n \leftarrow \bar{x}_n \odot \bar{B}, \\
\text{label generation:} \quad & \bar{w} \sim \mathcal{N}(0, I), \quad y_n \leftarrow \text{sign}(\bar{x}_n^\top \bar{w})
\end{aligned}$$

where $\odot$ is the element-wise multiplication. In the equations above, the first line describes a standard data sampling procedure from a multivariate Guassian distribution; the second line generates a magnitude vector $\bar{B}$, which is later sparsified by decreasing small elements that are smaller than a threshold $C_2$ with a factor of $C_1$; the third line describes the application of magnitude vectors on the dataset; and the fourth line generates a weight vector $\bar{w}$, and labels $y_n$, based on the signs of multiplications of data and the weights.

We should note that by the aforementioned data generation process, the parameters $C_1$ and $C_2$ control the sparsity of data points and the gradients: the smaller these two constants are, the sparser the gradients are; and the gradient of linear models on the dataset should be expected to be $\left((1 - C_2)d, C_2 \frac{C_1}{C_1+2}\right)$-approximately sparse. We set the dataset of size $N = 1024$, dimension $d = 2048$. In Figures 1, 2, 3, and 4, from the top row to the bottom row, the $\ell_2$ regularization parameter $\lambda$ is set to $1/(10N)$, $1/N$. And in each row, from the first column to the last column, $C_2$ is set to $4^{-1}$, $4^{-2}$, $4^{-3}$.

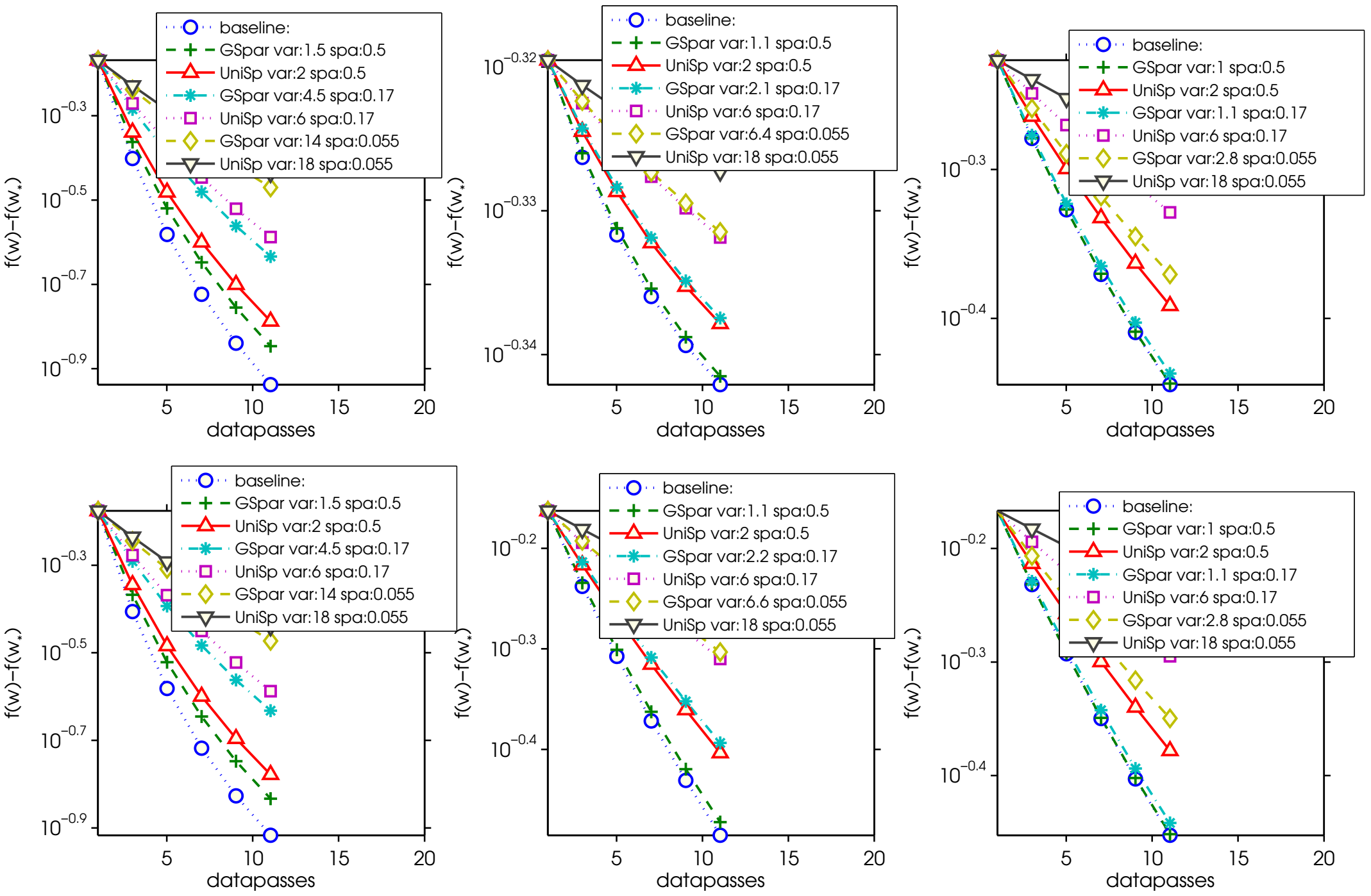

Figure 4: SVRG . Datasets generated by setting $C_1 = 0.9$. (Stronger sparsity)

The convergence trend measured by objective suboptimality is plotted in Figure 1, Figure 2, Figure 3, and Figure 4. Our algorithm is denoted by **GSpar**, and the uniform sampling method is denoted by **UniSp**, and the SGD/SVRG algorithm with non-sparsified communication is denoted by **baseline**, indicating the original distributed optimization algorithm. The $x$-axis shows the number of data passes, and the $y$-axis draws the suboptimality of the objective function $(f(w_t) - \min_w f(w))$. For the experiments, we report the sparsified-gradient SGD variance by the following rule

$$var := \frac{\sum_{t=1}^{T} \sum_{m=1}^{M} ||Q\left[g^m(w_t)\right]||^2}{\sum_{t=1}^{T} \sum_{m=1}^{M} ||g^m(w_t)||^2}$$

as the notation '$var$' in Figure 1 and Figure 2,, where $T$ is the iteration number when reaching a stopping point. The above formula indicates that the variance is measured by averaging over all $M$ workers. As for SVRG, similarly we have

$$var = \frac{\sum_{t=1}^{T} \sum_{m=1}^{M} ||Q\left[\nabla f(w_t) + g^m(w_t) - g^m(\widetilde{w})\right]||^2}{\sum_{t=1}^{T} \sum_{m=1}^{M} ||\nabla f(w_t) + g^m(w_t) - g^m(\widetilde{w})||^2}.$$

And '$spa$' in all figures represents the sparsity parameter $\rho$ in Algorithm 3. We observe that the theoretical complexity reduction against the baseline in terms of the communication rounds, which can be inferred by $var \times spa$, from the labels in Figures 1 to 4.

By comparing the results in Figure 1 and Figure 2, we observe that results on sparser data yields smaller gradient variance than results on denser data. Compared to uniform sampling, our algorithm generates gradients with less variance, and it converges much faster. This observation is consistent with the objective of our algorithm, which is to minimize gradient variance given a certain sparsity. The convergence slowed down linearly with respect to the increase of variance. The results on SVRG show better speed up — although our algorithm increases the variance of gradients, the convergence rate degrades only slightly.

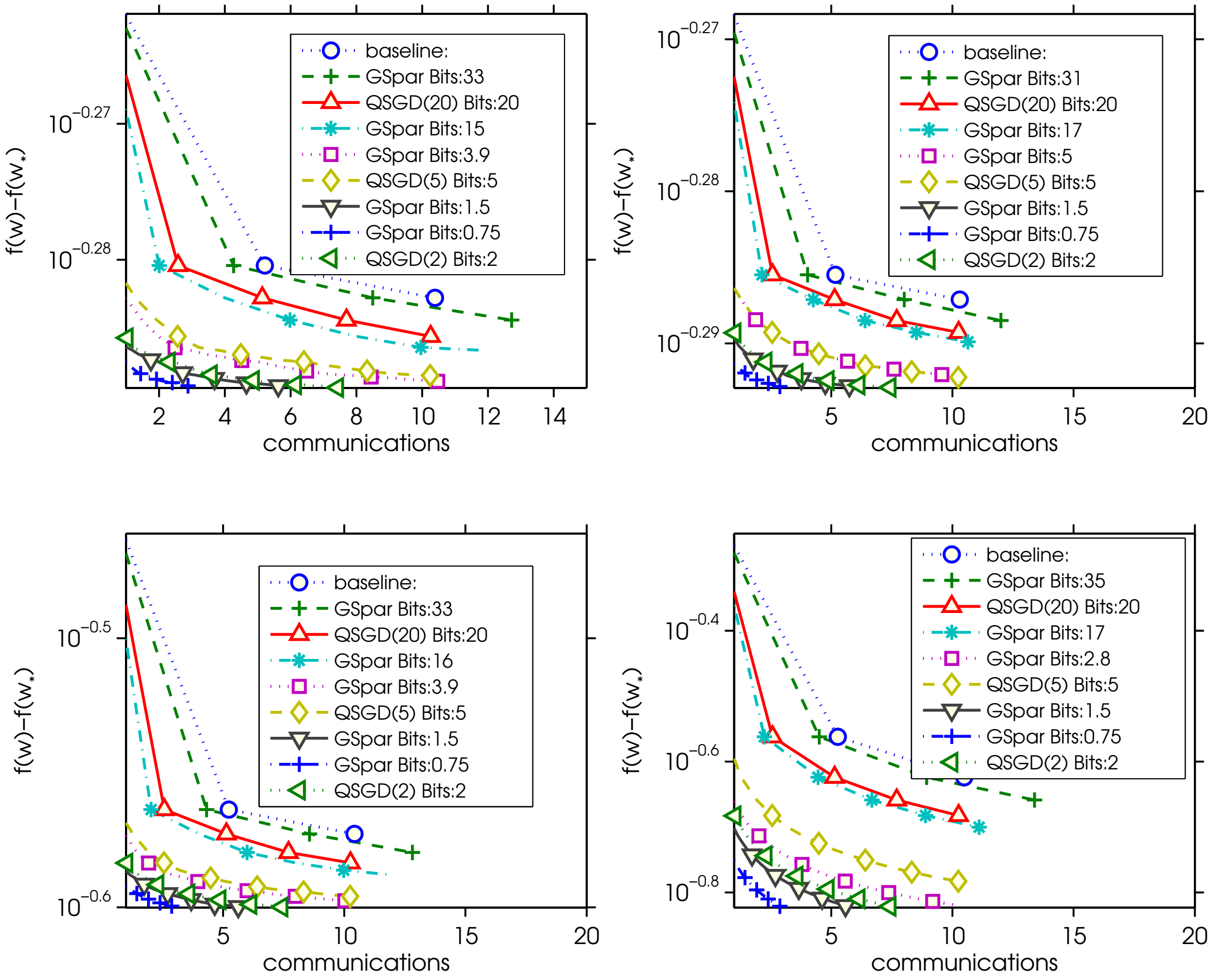

Figure 5: Comparison on SGD type approaches. Datasets generated by setting $C_1 = 0.6$. (Weaker sparsity)

We compared the gradient sparsification method with the quantized sparse gradient descent (QSGD) algorithm in [1]. For QSGD, elements of each stochastic gradient are quantized into $2^b$ discrete values (these discrete values can be represented using $b$ bits.):

$$Q(g_i, b) = \begin{cases} \mathrm{sign}(g_i)\lceil 2^b|g_i|\rceil 2^{-b}, & \text{if } \left(g_i - \mathrm{sign}(g_i)\lceil 2^b|g_i|\rceil 2^{-b}\right) 2^{-b} < Z_i \\ \mathrm{sign}(g_i)\lfloor 2^b|g_i|\rfloor 2^{-b}, & \text{if } \left(g_i - \mathrm{sign}(g_i)\lceil 2^b|g_i|\rceil 2^{-b}\right) 2^{-b} \geq Z_i. \end{cases}$$

where $Z_i$ is sampled from a uniform distribution in $[0, 1]$. The results are shown in Figures 5 and 6. The data are generated as previous, with both strong and weak sparsity settings. From the top row to the bottom row, the $\ell_2$ regularization parameter $\lambda$ is set to $1/(10N)$, $1/N$. And in each row, from the first column to the last column, $C_2$ is set to $4^{-1}$, $4^{-2}$. In this comparison, we use the overall communication coding length of each algorithm, and note the length in x-axis. For QSGD, the communication cost per element is

$$\mathcal{H}(T, M) := TMb.$$

where $b$ refers to the bits of floating point number. QSGD($b$) denotes QSGD algorithm with bit number $b$ in these figures, and the average bits required to represent per element is on the labels. For gradient sparsification, the communication cost per element is calculated as

$$\mathcal{H}(T, M) := \sum_{t=1}^{T} \sum_{m=1}^{M} \left( \sum_i \mathbb{I}(p_i^{t,m} = 1)(b + \log_2 d) + \min \left( 2d, \log_2 d \sum_i \mathbb{I}(p_i^{t,m} < 1) p_i \right) + b \right),$$

where $\mathcal{H}(T, M)$ refers to the communication cost before the $T$-th iteration, and $p^{t,m}$ refers to the probability vector of the gradient calculated by the $m$-th worker during the $t$-th iteration, and the illustration to the formulation can be found in the previous section. We did not apply aggressive gradient compression in both algorithms, like very low-bit formats or very sparse representation, which will lead to too many iterations of calculation to be practical. And since the second order momentum of gradients of QSGD is hard to explicitly calculated, we set the step size of both algorithms to be irrelevant with gradient variance, as $\eta_t \propto 1./t$. From Figures 5 and 6, we observe that the proposed sparsification approach is at least comparable to QSGD, and significantly outperforms QSGD when the gradient sparsity is stronger; and this concords with our analysis on the gradient approximate sparsity encouraging faster speed up.There is still improvement space in reducing the floating point precisions for gradient sparsification, which is beyond the paper.

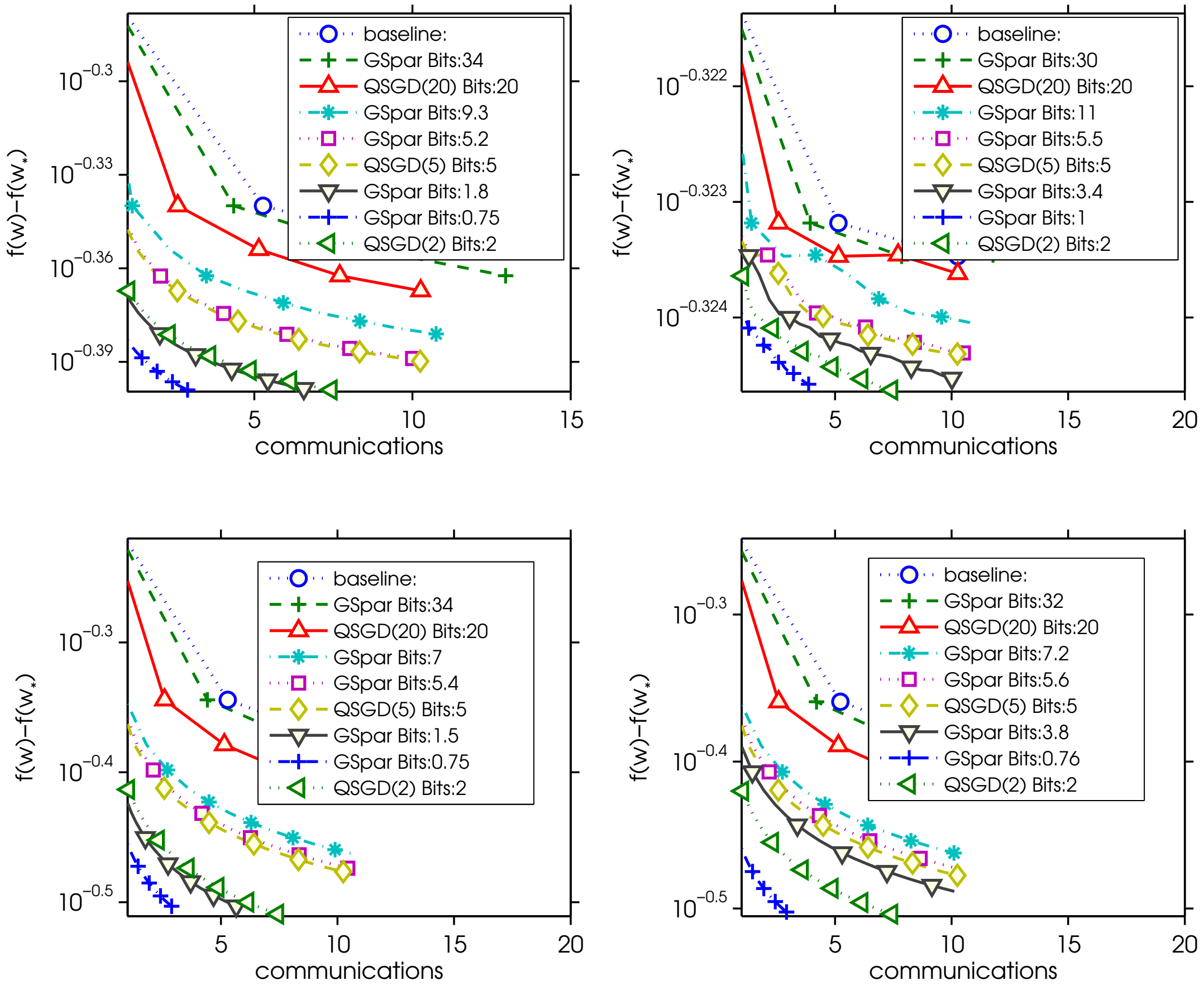

Figure 6: SGD . Datasets generated by setting $C_1 = 0.9$. (Stronger sparsity)

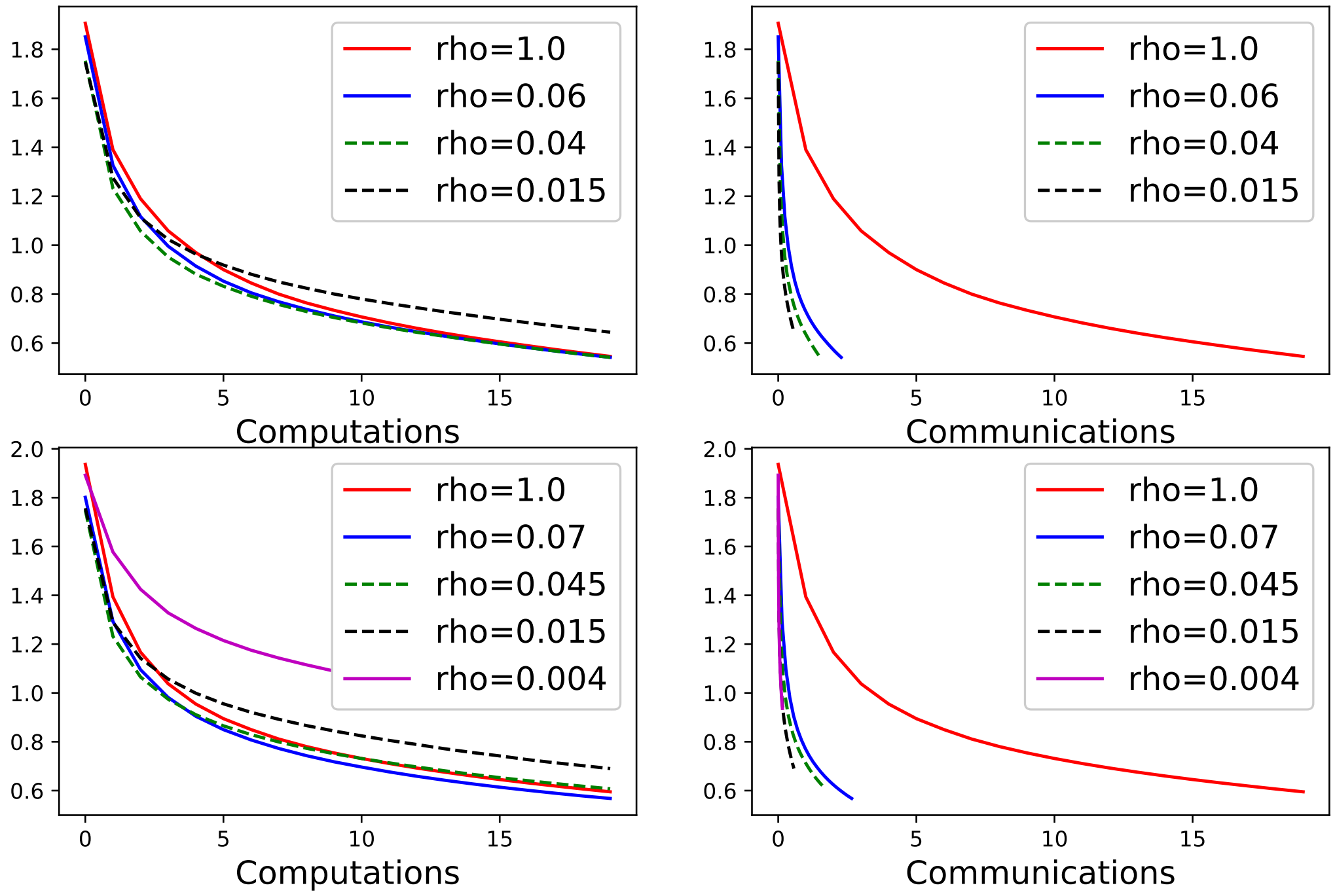

Figure 7: Comparison of convolutional neural networks of 3 layers of channels of 32 (top) and 24 (bottom) on CIFAR10. (Y-axis: loss function $f(w_t)$.)

## 5.2 Experiments on deep learning

This section conducts experiments on non-convex problems. we consider the convolutional neural networks on the CIFAR10[1] dataset. We implement our method using the neon framework[2] provided by Intel Nervana System. The convolution operator is accelerated by winograd algorithm [11]. We experiment with neural networks using different settings. generally, the network consists of three convolutional layers $(3 \times 3)$, two pooling layers $(2 \times 2)$, and one 256 dimensional fully connected layer. Each convolution layer is followed by a batch-normalization layer. the channels of each convolutional layer is set to $\{24, 32, 48, 64\}$. We use the ADAM optimization algorithm [9], and the initial step size is set to 0.02. since there exist major differences among weight magnitudes of different network layers, the sparsification is done independently over each layer.

[1] https://www.cs.toronto.edu/ kriz/cifar.html
[2] https://github.com/nervanasystems/neon

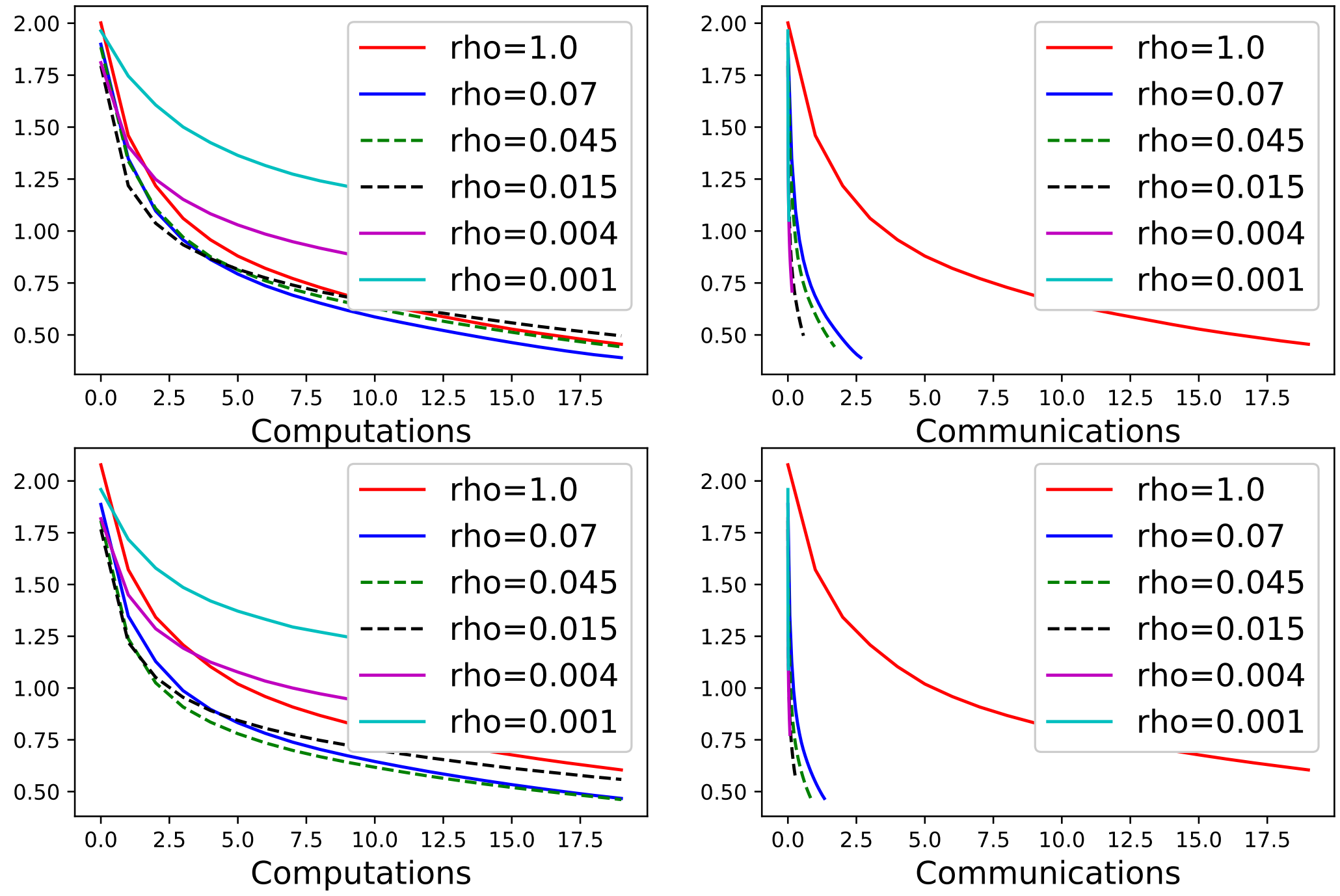

Figure 8: Convolutional neural networks of 3 layers of channels of 64 (top) and 48 (bottom) on CIFAR10. (Y-axis: objective function $f(w_t)$.)

In Figure 7 and Figure 8, we plot the objective loss against the computational complexity measured by the number of epochs (1 epoch is equal to 1 pass of all training samples). We also plot the convergence with respect to the communication cost, which is the product of computations and the sparsification parameter $\rho$. The experiments on each setting are repeated 4 times and we report the average objective function values. The results show that for this non-convex problem, the gradient sparsification slows down the training efficiency only slightly. In particular, the optimization algorithm converges even when the sparsity ratio is about $\rho = 0.004$, and the communication cost is significantly reduced in this setting. This experiments also shows that the optimization of neural networks are less sensitive to gradient noise, and the noises within a certain range may even help the algorithm to avoid being trapped in bad local minimal.

### 5.3 Experiments on asynchronous parallel SGD

In this section, we study parallel implementations of SGD on the multi-core architecture. A number of bench-mark algorithms, including hogwild! [17, 5], ASGD [15], and ASCD [16] are considered for comparison. There are three types of update schemes: *Lock*, *Atomic*, and *Wild*. During the update of the *Lock* scheme, vector coordinates are locked so that there is only one thread reading and writing the memory. However, the running time of this method is the slowest due to the frequent lock conflicts. The *Atomic* scheme is a trade-off between consistency and running time, where each coordinate is atomically updated, but different coordinates may be simultaneously updated by different threads. The *Wild* scheme is a lock-free approach, where all the threads simultaneously update all memory locations, causing writing conflicts, which potentially lead to multiple invalid updates. Although this scheme has the weakest consistency, empirically it runs the fastest without loss of training accuracy.

**Algorithm 4** An asynchronous parallel optimization algorithm

1: Initialize $t_m = 0$ for all worker $m$, initialize the weight $w$.
2: **repeat**
3: Each worker updates its local clock $t_m = t_m + 1$ and calculates local stepsize $\eta_{t_m}$
4: Each worker $m$ calculates $g^m(w_t)$ based on local data.
5: Calculate the probability vector $p^m$ by Algorithm 3.
6: Sparsify the gradients to $Q(g^m(w_t))$.
7: Sequentially modifying each coordinate of the global weight vector $w_i = w_i - \frac{1}{M}\eta_{t_m} Q(g^m(w_t))_i$ by using atomic operations.
8: **until** convergence

We employ the support vector machine for binary classification, where the loss function is

$$f(w) = \frac{1}{N}\sum_n \max(1 - a_n^\top w b_n, 0) + \lambda_2 ||w||_2^2, \quad a_n \in \mathbb{R}^d, \quad b_n \in \{-1, 1\}. \tag{16}$$

We implemented shared memory multi-thread SGD, where each thread employs a locked read, which may block other threads' writing to the same coordinate. We use atomic instructions for updating coordinates.

To improve the speed of the algorithm, we also employ several engineering tricks. First, we observe that $\forall p_i < 1, \quad g_i/p_i = \text{sign}(g_i)/\lambda$ from (5); therefore we only need to assign constant values to these variables, without applying float-point division operations. Another costly operation is the pseudo-random number generation in the sampling procedure; therefore we generate a large array of pseudo-random numbers in $[0, 1]$, and iteratively read the numbers during training without calling a random number generating function.

The data are generated as following.

$$\begin{aligned} \text{dense data generation:} \quad & \bar{x}_{ni} \sim \mathcal{N}(0,1), \quad \forall i \in [d], n \in [N], \quad \bar{w} \sim \text{Uniform}[-0.5, 0.5]^d, \\ \text{data sparsification:} \quad & \bar{B} \sim \text{Uniform}[0,1]^d, \quad \bar{B}_i \leftarrow C_1 \bar{B}_i, \quad \text{if:} \bar{B}_i \le C_2, \quad \forall i \in [d], \quad x_n \leftarrow \bar{x}_n \odot \bar{B}, \\ \text{label generation:} \quad & y_n \leftarrow \text{sign}(x_n^\top \bar{w} + \sigma), \quad \text{where} \quad \sigma \sim \mathcal{N}(0,1) \end{aligned}$$

In the equations above, the first line describes a standard data sampling procedure from a multivariate Guassian distribution, and a sampled weight vector $\bar{w}$ from multivariate uniform distributions; the second line sparsifies the dataset by a sparse magnitude vector $\bar{B}$; the third line generates a weight vector $\bar{w}$, and labels $y_n$, based on the signs of multiplications of data and the weights, plus noises. We set the dataset of size $N = 51200$, dimension $d = 256$, also set $C_1 = 0.01$ and $C_2 = 0.9$. We train $\ell_2$ regularized support vector machines, where the regularization parameter $\lambda_2$ is denoted by $reg$, and the number of threads is denoted by $workers$. In practice, we also set the initial step size of each algorithm to $lrt/\rho$. The number of workers is set to 16 or 32, the regularization parameter is set to $\{0.5, 0.1, 0.05\}$, and the learning rate is chosen from $\{0.5, 0.25, 0.05, 0.25\}$. The convergence of objective value against running time (milliseconds) is plotted in Figure 9.

From Figure 9, we can observe that using gradient sparsification, the conflicts of multiple threads for reading and writing the same coordinate are significantly reduced. Therefore the training speed is significantly faster. One can also observe that the sparsification technique works better at 32 threads than 16, since the more threads, the more frequently the lock conflicts occur.

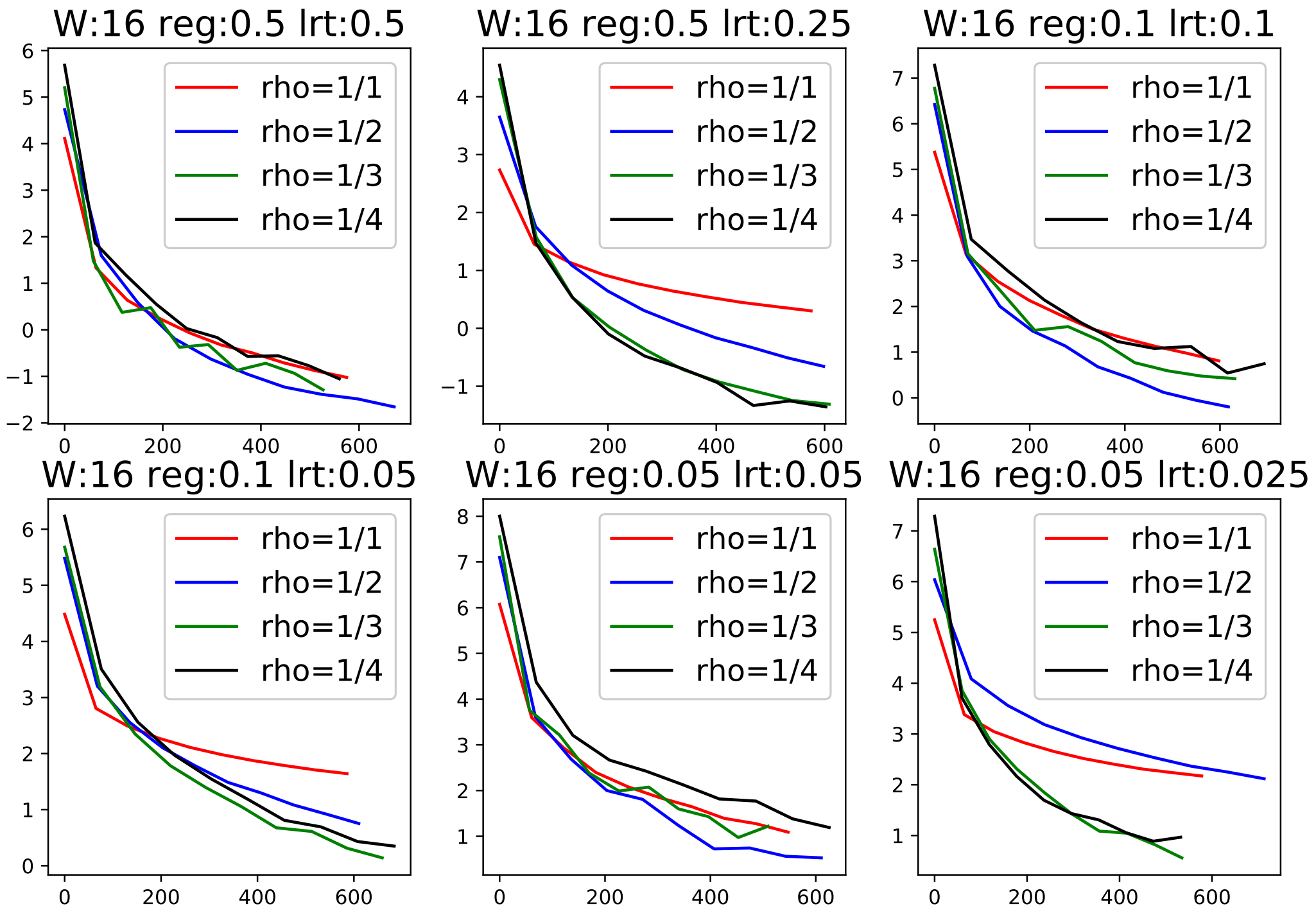

Figure 9: Loss functions by multi-thread atomically updating SVM. X-axis: time in milliseconds, Y-axis: the loss function in logarithm $\log_2(f(w_t))$. (Including above and below figures)

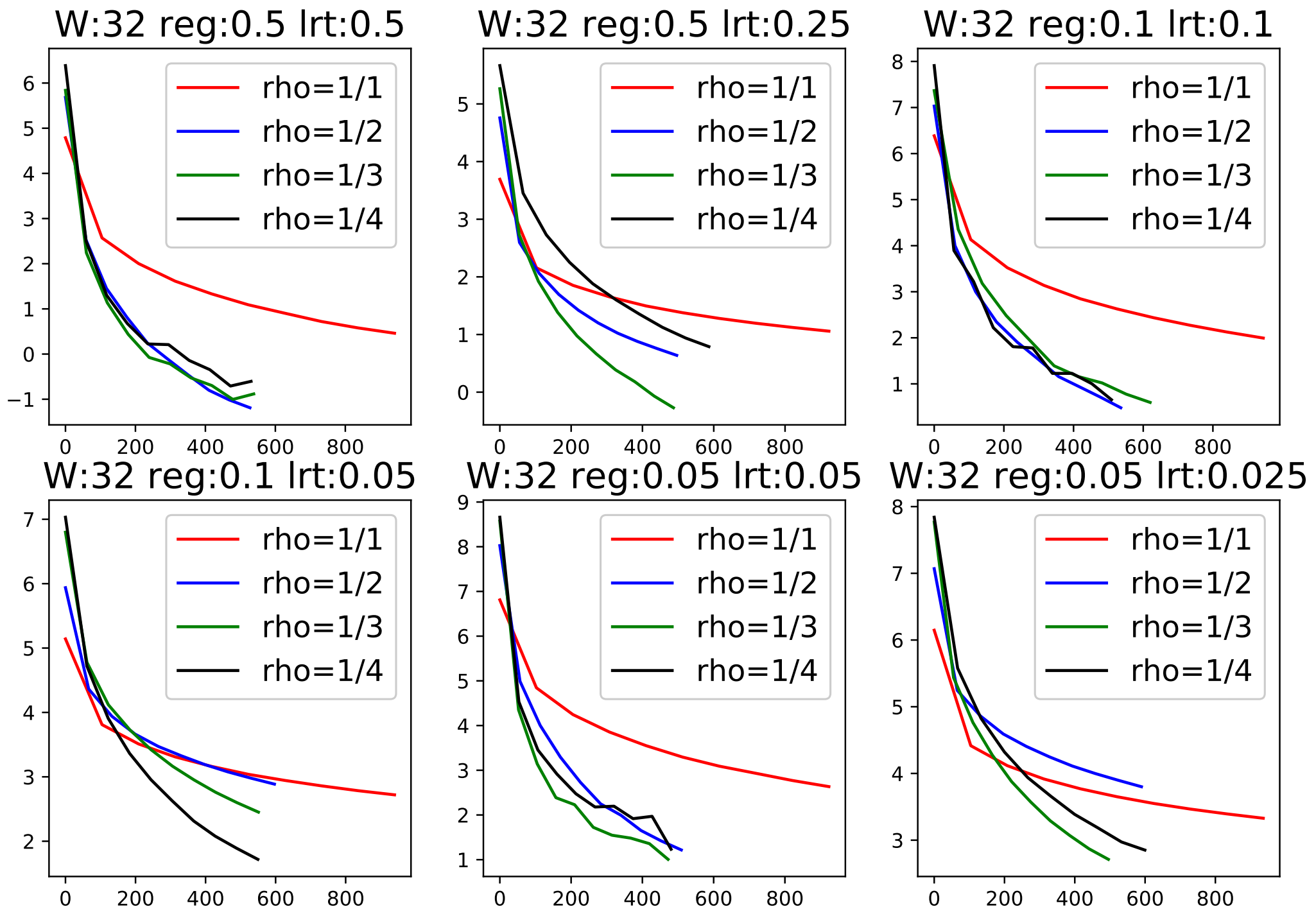

# 6 Conclusions

In this paper, we propose a gradient sparsification technique to reduce the communication cost for large scale distributed machine learning. The key idea is to randomly drop out coordinates of the stochastic gradient vectors and amplify the remaining coordinates appropriately to ensure the sparsified gradient to be unbiased. We propose a convex optimization formulation to minimize the coding length of stochastic gradients given the variance budget that monotonically depends on the computational complexity. To solve the optimal sparsification efficiently, several simple and fast algorithms are proposed for approximate solutions, with theoretically guaranteed sparsity. The experiments on $\ell_2$ regularized logistic regression and convolutional neural networks show that the proposed sparsification technique can effectively reduce the communication cost during training. Moreover, the algorithm can also be used for shared memory architectures, and experiments on multi-thread SVM showed that our method significantly improved the running time by reducing conflicts among multiple threads when they competed for the same shared memory resources.